\documentclass[runningheads]{llncs}
\usepackage{graphicx}
\usepackage{tikz}
\usetikzlibrary{arrows,automata,calc,decorations.pathreplacing,positioning,shapes}
\usepackage{pgfplots}
\usepackage{subcaption}
\usepackage[frozencache,cachedir=minted-cache]{minted}
\usepackage{listings}
\usepackage{hyperref}
\newenvironment{code}{\captionsetup{type=listing}}{}

\tikzset{%
  arrow/.style={thick,-stealth},
  edge/.style={midway,sloped,above},
  entity/.style={circle,draw},
  label/.style={yshift=0.2cm},
}

\definecolor{blueLink}{HTML}{180CAD}
\definecolor{dark}{HTML}{7B7D7B}
\definecolor{grey}{HTML}{C6C7C6}
\definecolor{green}{HTML}{789437}
\definecolor{blue}{HTML}{377894}
\definecolor{grey}{HTML}{fef9f6}
\definecolor{red}{HTML}{943778}
\definecolor{mygreen}{HTML}{bdd7d6}
\definecolor{myred}{HTML}{dec3d6}

\definecolor{myyellow}{HTML}{d6dec3}
\definecolor{mygreen}{HTML}{c3ded9}
\definecolor{myblue}{HTML}{c3d6de}
\definecolor{darkBlue}{HTML}{5f8ba8}
\definecolor{mypurple}{HTML}{c3c9de}
\definecolor{darkPurple}{HTML}{485684}
\definecolor{myyellow}{HTML}{DED8C3}
\definecolor{mybrown}{HTML}{DEC3C9}
\definecolor{darkRed}{HTML}{844871}
\begin{document}
\title{\texttt{pyRDF2Vec}: A Python Implementation and Extension of RDF2Vec}
%
%
\author{Gilles Vandewiele \and Bram Steenwinckel \and Terencio Agozzino \and Femke Ongenae}
%
\authorrunning{G. Vandewiele et al.}
%
\institute{IDLab, Ghent University -- imec, 9000 Gent, Belgium}
\maketitle              
\begin{abstract}
This paper introduces \texttt{pyRDF2Vec}\footnote{\url{https://github.com/IBCNServices/pyRDF2Vec}}, a Python software package that reimplements the well-known RDF2Vec algorithm along with several of its extensions. By making the algorithm available in the most popular data science language, and by bundling all extensions into a single place, the use of RDF2Vec is simplified for data scientists. The package is released under a MIT license and structured in such a way to foster further research into sampling, walking, and embedding strategies, which are vital components of the RDF2Vec algorithm. Several optimisations have been implemented in \texttt{pyRDF2Vec} that allow for more efficient walk extraction than the original algorithm. Furthermore, best practices in terms of code styling, testing, and documentation were applied such that the package is future-proof as well as to facilitate external contributions.

\end{abstract}

\section{Introduction}
\setcounter{footnote}{0} 
Knowledge Graphs (KGs) are an ideal candidate to perform hybrid Machine Learning (ML) where both background and observational knowledge are taken into account to construct predictive models. However, since KGs are symbolic data structures, they cannot be fed to ML algorithms directly and first require a non-trivial transformation step in which symbolic substructures of the graph are converted into numerical representations. These transformation techniques can typically be classified as being \emph{feature}-based or \emph{embedding}-based~\cite{wang2017knowledge}. Feature-based approaches are often interpretable, but require domain knowledge about the task at hand and are effort-intensive. Embedding-based approaches, on the other hand, are typically agnostic to the task and are usually able to outperform their feature-based counterparts. Resource Description Framework To Vector (RDF2Vec)~\cite{ristoski2019rdf2vec} is an unsupervised, task-agnostic, and embedding-based approach that has gained significant popularity over the past few years. RDF2Vec builds on the popular Natural Language Processing (NLP) technique Word2Vec. The latter generates embeddings for different tokens present in a corpus, by training a neural network in an unsupervised way that must predict either a token based on its context (Continuous Bag of Words) or the context based on a token (Skip-Gram). The corpus, fed to Word2Vec, is constructed by extracting a large number of walks from the KG. A walk is a sequence of entities obtained from the KG by starting at a certain entity and traversing the directed edges. \\

\begin{figure}[h!]
    \centering
    \includegraphics[width=0.6\textwidth]{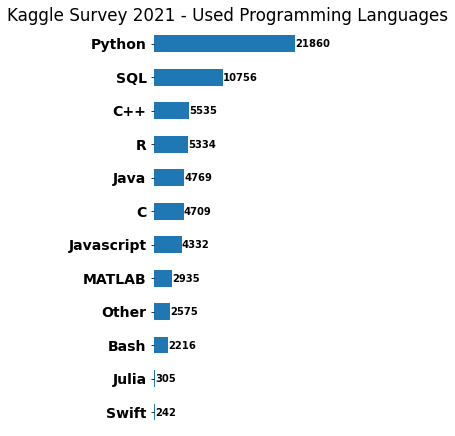}
    \caption{Programming languages used by data scientists according to the Kaggle Survey 2021.}
    \label{fig:lang_popularity}
\end{figure}

Since its initial publication, in 2017, many extensions to the algorithm have been proposed. However, each of these extensions are individual implementations, which complicates combining several of them. Moreover, the original code for RDF2Vec was written in Java, which is significantly less popular than Python for data science, according to the Kaggle Survey 2021\footnote{\url{https://www.kaggle.com/c/kaggle-survey-2021}}. In Figure~\ref{fig:lang_popularity}, the answers to the question ``What programming languages do you use on a regular basis?'', where multiple answers were possible, are depicted. It should be noted that among the 4769 people who selected Java as being used regularly, only 598 did not pick Python. This makes it difficult to integrate the original RDF2Vec implementation into a data science pipeline, which is typically written in Python.

In this paper, we present \texttt{pyRDF2Vec}, a Python implementation of the original algorithm and many of its extensions. Moreover, various mechanisms are built in which allows to better handle large KGs. The code is released under an open-source license and is written in a way to facilitate further research into the different components of the RDF2Vec algorithm. The remainder of this paper is structured as follows. In Section~\ref{sec:background}, we provide background on representation learning for KGs, followed by an in-depth discussion of RDF2Vec and its extensions. Then, in Section~\ref{sec:pyrdf2vec}, we present the architecture of our \texttt{pyRDF2Vec} package and the mechanisms set in-place to easily allow for contributions by others. In Section~\ref{sec:usage}, we discuss some studies and other software packages that have already made use of \texttt{pyRDF2Vec}. Finally, we conclude our paper in Section~\ref{sec:conclusion}. In Appendix~\ref{appendix}, we provide a code snippet that shows how \texttt{pyRDF2Vec} can be used. 

\section{Background}\label{sec:background}

In this section, we describe the necessary background to elaborate upon \texttt{pyRDF2Vec}. First, we will discuss related work regarding the transformation of a KG into numerical representations. Afterwards, we outline an in-depth overview of how RDF2Vec works and its extensions released over the past few years.

\subsection{Representation Learning}

As mentioned in the introduction, a \emph{feature}-based or \emph{embedding}-based transformation step is required that converts the symbolic KGs into numerical vectors before they can be used in ML models. Especially embedding-based approaches, which make use of Deep Learning techniques, have gained increasing popularity over the past few years as these can be applied out-of-the-box and can run efficiently on Graphical Processing Units (GPUs), which are quite commonly available today. Moreover, the largest advantage of embedding-based techniques is that they are typically task-agnostic and as such do not require extensive domain knowledge and/or significant effort, as opposed to feature-based approaches. A further distinction can be made between embedding-based techniques. A first category consists of techniques that learn embeddings either through tensor factorisation or through negative sampling~\cite{nickel2015review,choudhary2021survey,wang2017knowledge}, e.g. TransE~\cite{bordes2013translating}. A second category consists of Deep Learning architectures that make use of parameterised transformations, based on information from the neighbourhood of a node that is collected through message passing~\cite{schlichtkrull2018modeling}, e.g. Relational Graph Convolutional Networks (R-GCN). The parameters of this transformation are learned through back-propagation in a supervised fashion. A third, and final, category adapts existing NLP techniques, such as Word2Vec~\cite{mikolov2013efficient}, to work on graph structures. RDF2Vec belongs to this final category~\cite{ristoski2019rdf2vec}.

\subsection{RDF2Vec}
RDF2Vec is an unsupervised, task-agnostic algorithm that achieves state-of-the-art performances on many benchmark datasets~\cite{ristoski2019rdf2vec}. It extends Word2Vec to work on graph structures by first extracting walks that serve as corpus. Each walk can be seen as a sentence of a corpus and each hop within such walks corresponds to a token. Word2Vec will then learn embeddings for each of these tokens in an unsupervised matter by learning to predict either a token based on its context (Continuous Bag of Words), or the context based on a token (Skip-Gram). Over the past few years, several extensions to RDF2Vec have been suggested, which we will discuss subsequently. A good up-to-date overview on how RDF2Vec works, which extensions have been proposed over the last few years, and of applications that make use of RDF2Vec can be found on a website hosted by the original authors\footnote{\url{www.rdf2vec.org}}.\\

The number of walks that can be extracted quickly grows, depending on the depth of those walks and the size of the KG. As such, exhaustively extracting every possible walk becomes infeasible rather quickly. As a solution, Cochez et al.~\cite{cochez2017biased} proposed several sampling, or biased walking, techniques which enable to only extract a subset of walks that still capture most of the information. Recently, more sampling strategies have been proposed: (i) utilising page transition probabilities~\cite{taweel2020towards}, (ii) using Metropolis-Hastings sampling~\cite{zhang2021discovering}, or (iii) other forms of prior knowledge~\cite{mukherjee2019graph}. \\

Originally, the RDF2Vec algorithm used random walking and the Weisfeiler-Lehman paradigm to extract the corpus of walks for Word2Vec. However, within the domain of graph-based ML, walking techniques that are more advanced than random sampling have been suggested over the past few years. In addition, it has been shown that the Weisfeiler-Lehman paradigm introduces little to no extra information in the extracted walks. As such, Vandewiele et al. evaluated different walking strategies on several benchmark datasets to show that there is no one-size-fits-all strategy, and that tuning the strategy for the task at hand can result in increased performances~\cite{vandewiele2020walk}. \\

Finally, Portisch et al.~\cite{portisch2021putting} applied an order-aware variant of Word2Vec to the corpus extracted by the walking and sampling strategies, which resulted in significant increased predictive performances on multiple benchmark datasets.

\section{\texttt{pyRDF2Vec}}\label{sec:pyrdf2vec}

In this section, we elaborate upon our \texttt{pyRDF2Vec} package. We first present its architecture, then give an overview of all the extensions available today and finally discuss the different mechanisms implemented to facilitate external contributions.

\subsection{Architecture}

\begin{figure}[!ht]
  \centering
  \resizebox{\linewidth}{!} {
    \begin{tikzpicture}[>=stealth',
      extract_walks/.style={draw,minimum width=.8cm,minimum height=.8cm,fill=mygreen!40},
      process/.style={draw,minimum width=1cm,minimum height=1cm,node distance=0.35cm,fill=mybrown!20}
      ]
    \node[draw,minimum width=2.5cm,minimum height=3cm,loosely dashed,color=darkBlue] at (0,0) (graph_entities) {};
    \node[draw,minimum width=2cm,minimum height=1cm,yshift=-65pt,above=of graph_entities,fill=myblue] (graph) {Graph};
    \node[draw,minimum width=2cm,minimum height=1cm,yshift=15pt,below=of graph,fill=myblue] (entities) {Entities};
    \node[draw,minimum width=2cm,minimum height=1cm,yshift=2pt,above=of graph,fill=myblue!50] (connector) {Connector};

    \node[inner sep=0pt,xshift=-20pt,above=of connector] (rdf) {\includegraphics[width=0.074\textwidth]{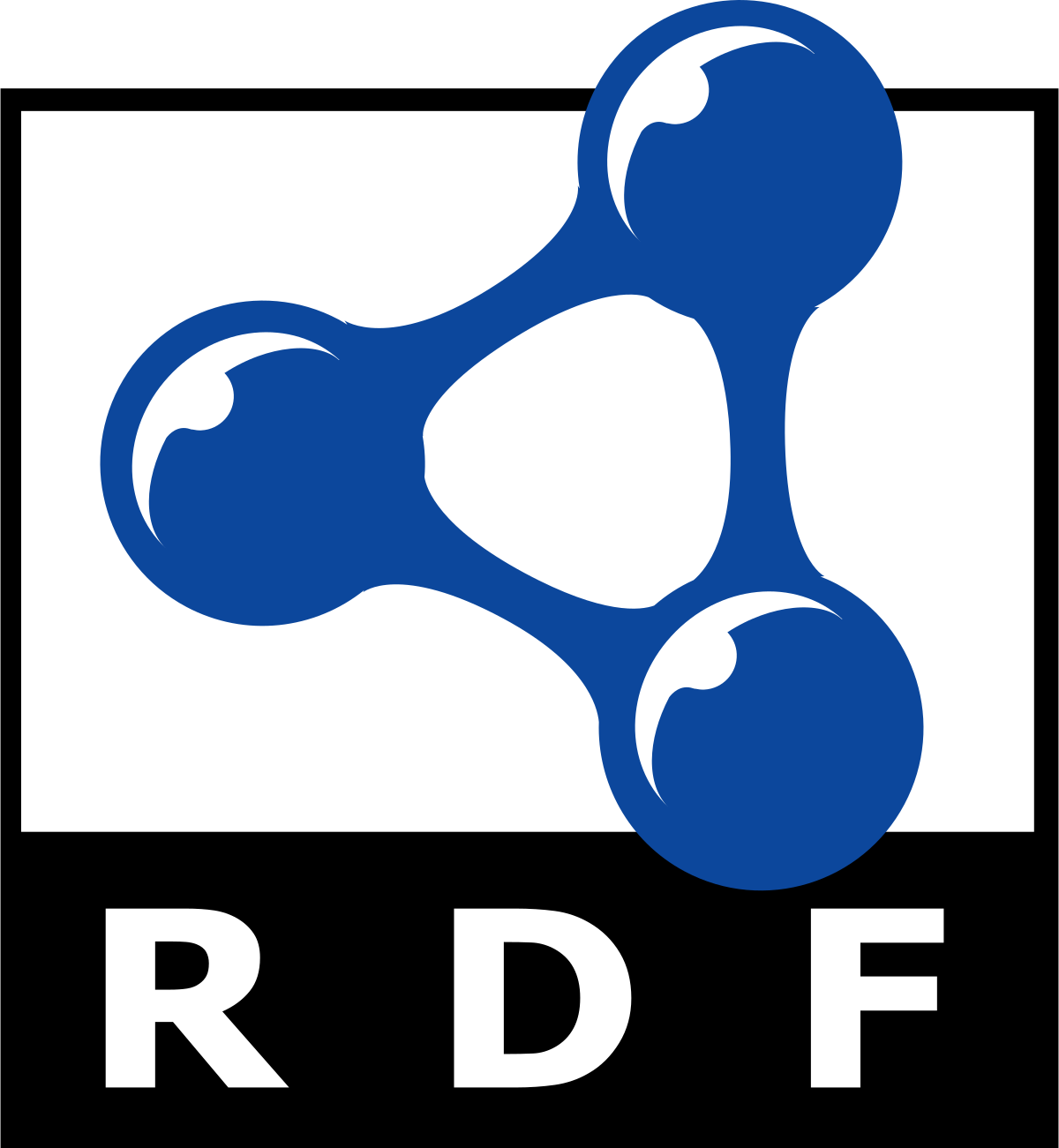}};
    \node[inner sep=0pt,xshift=20pt,above=of connector] (sparql) {\includegraphics[width=0.08\textwidth]{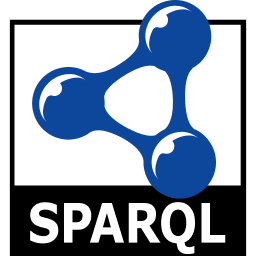}};

    \node[label,below of=graph_entities,yshift=-1.1cm,color=darkBlue] {\textbf{Inputs}};

    \node[draw,minimum width=3cm,minimum height=1cm,right=of graph_entities,fill=myyellow] (transformer) {Transformer};
    \node[draw,minimum width=2.5cm,minimum height=3cm,loosely dashed,color=darkRed,right=of transformer] (walker_sampler) {};
    \node[draw,minimum width=2cm,minimum height=1cm,yshift=-65pt,above=of walker_sampler,fill=myred] (walker) {Walker};
    \node[draw,minimum width=2cm,minimum height=1cm,yshift=15pt,below=of walker,fill=myred] (sampler) {Sampler};
    \node[draw,minimum width=3cm,minimum height=1cm,above=of transformer,fill=mypurple!50] (embedder) {Embedder};
    \node[draw,minimum width=3cm,minimum height=1cm,above=of embedder,fill=mypurple] (embeddings) {Embeddings};
    \node[draw,minimum width=3cm,minimum height=1cm,node distance=0.5cm,right=of embeddings,fill=mypurple] (literals) {Literals};

    \node[label,above of=walker_sampler,xshift=0.3cm,yshift=20pt,color=darkRed] {\textbf{Strategy}};

    \node[draw,minimum width=7.5cm,minimum height=1.5cm,xshift=1.8cm,yshift=-2.3cm,above=of embeddings,loosely dashed,color=darkPurple] (output) {};
    \node[label,above of=output,xshift=2.5cm,yshift=-0.1cm,color=darkPurple] {\textbf{Outputs}};
    
    \node[inner sep=0pt,xshift=2cm, right=of output] (sklearn) {\includegraphics[width=0.15\textwidth]{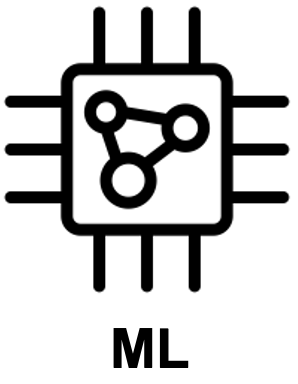}};

    \node[draw,minimum width=3cm,minimum height=1cm, node distance=2.3cm,right=of walker,fill=mygreen] (walks) {Walks};

    \draw[arrow,shorten >=0.2cm,shorten <=0.2cm] (transformer) -- (walker_sampler) node[midway,above] {(2)};

    \draw[arrow,shorten >=0.2cm,shorten <=0.2cm] (rdf.south) -- ([xshift=-20pt]connector.north);
    \draw[arrow,shorten >=0.2cm,shorten <=0.2cm] (sparql.south) -- ([xshift=20pt]connector.north);
    \draw[arrow,shorten >=0.05cm,shorten <=0.2cm] (connector) -- (graph);



     \draw[arrow,shorten >=0.1cm,shorten <=0.1cm] (walker.east) -- (walks.west);


    \draw[arrow,shorten >=0.1cm,shorten <=0.1cm] (sampler) -- (walker);
    \draw[arrow,shorten >=0.2cm,shorten <=0.2cm] (transformer) -- (embedder) node[midway,right] {(3)};
    \draw[arrow,shorten >=0.05cm,shorten <=0.2cm] (embedder) -- (embeddings);
    \draw[arrow,shorten >=0.2cm,shorten <=0.2cm] (output) -- (sklearn);

    \draw[arrow,shorten >=0.2cm,shorten <=0.2cm] (graph_entities) -- (transformer.west) node[midway,above] {(1)};

    \draw[shorten <=0.2cm] ([xshift=0.2cm]transformer) -- ([yshift=1.3cm]walker_sampler.north);
    \draw[arrow,shorten >=0.05cm] ([yshift=1.3cm]walker_sampler.north) -- (literals);

    \draw ([xshift=10pt]walks.east) |- ([yshift=-20pt]sampler.south);
    \draw[shorten <=0.1cm] (walks) -- ([xshift=10pt]walks.east);
    \draw[arrow,shorten >=0.2cm] ([yshift=-20pt]sampler.south) -| (transformer.south);
  \end{tikzpicture}
  }
  \caption{Workflow of \texttt{pyRDF2Vec}. A \texttt{Graph} and collection of \texttt{Entities} are provided by the user to the \texttt{Transformer} (1), which is instantiated with a list of different strategies consisting of a \texttt{Walker} and \texttt{Sampler} (2). The latter are responsible for extracting walks from the \texttt{Graph} which are, in turn, fed to the embedder to calculate \texttt{Embeddings} (3). In addition, the \texttt{Transformer} also extracts \texttt{Literals} by following paths specified by the user. }
  \label{tikz:workflow}
\end{figure}
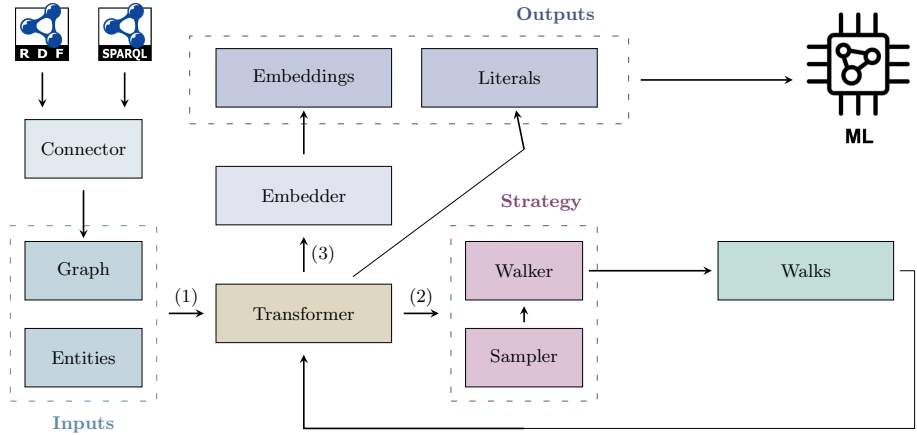

In Figure \ref{tikz:workflow}, an overview of the \texttt{pyRDF2Vec} workflow is provided. Seven main modules are used, which we now discuss subsequently.

\begin{enumerate}
\item \textbf{Connector}: coordinates the interaction with a local or remote graph. For KGs located on hard disk, \texttt{pyRDF2Vec} uses \texttt{rdflib} to load the graph into memory. If required, walk extraction from remote graphs is also possible through a SPARQL endpoint. Additional connectors can be implemented based on the provided \texttt{Connector} base class.
\item \textbf{Graph}: is the internal representation of the KG. It is used to efficiently traverse the graph and to store additional information regarding nodes and edges.
\item \textbf{Entities}: is the set of nodes within the graph for which we want to generate embeddings. These entities will serve as the starting points for the walk extraction and need to be provided by the user. It should however be noted that in fact all of the entities that appear in these extracted walks will have an associated embedding.
\item \textbf{Transformer}: the main interface for users that combines all other components.
\item \textbf{Sampler}: prioritises the use of some edges in the graph over others using a weight allocation strategy. The current \texttt{pyRDF2Vec} version implemented each of the sampling techniques described by Cochez et al.~\cite{cochez2017biased}. Additional sampling techniques can easily be implemented, according to the provided \texttt{Sampler} base class. 
\item \textbf{Walker}: responsible for extracting walks from the KG. Different walking strategies, proposed by Vandewiele et al.~\cite{vandewiele2020walk} are incorporated in the current \texttt{pyRDF2Vec} version. New walking strategies can be implemented using the \texttt{Walker} base class.
\item \textbf{Embedder}: is in charge of transforming the extracted walks into embeddings, based on a trained model. By default, Word2Vec is used within this embedder code to generate these embeddings. A fastText~\cite{grave2018learning} embedder is also made available in the current \texttt{pyRDF2Vec} version and additional embedding techniques can be added by using the \texttt{Embedder} base class.
\end{enumerate}

It is important to \texttt{Connector}, \texttt{Sampler}, \texttt{Walker}, and \texttt{Embedder} expose interfaces that can be implemented by users. That way, we hope to both facilitate and stimulate further research into these components of the RDF2Vec algorithm.

\subsection{Optimizations and extensions}

The \texttt{pyRDF2Vec} implementation has several extensions, that speed up walk extraction and which provide information in addition to the embeddings based on walks. \\

First, the \texttt{Transformer} takes a list of \texttt{Walker} strategies, with optionally associated \texttt{Sampler} strategies, which enables to combine several strategies. This allows for further research into techniques similar to ensembling, where the information obtained from several strategies is combined. This combination can be done either (i) on corpus-level, by concatenating the walks extracted by the different strategies together before feeding them to the \texttt{Embedder}, (ii) on embedding level, where embeddings are learned on the corpora of each strategy individually and then aggregated, or (iii) on prediction level, where the embeddings learned on each corpora are fed to a classifier to make predictions for the downstream task and then aggregated. The combination of different strategies is illustrated in the example code provided in Appendix~\ref{appendix}. \\

A second extension in the \texttt{pyRDF2Vec} allows to extract literal information in addition to the embeddings learned, based on the graph structure surrounding entities of interest. To achieve this, the user can specify a set of paths, starting from the nodes provided in \texttt{Entities}, for which literal information can be found. \texttt{pyRDF2Vec} will then traverse these paths and return (i) \texttt{NaN} if the literal cannot be found, (ii) a scalar in case exactly one literal can be found, and (iii) a list of literals in case the path to a literal can be found multiple times. From then on, the user can process this information and concatenate this to the provided embeddings. \\

\texttt{pyRDF2Vec} enables reverse walking by traversing across incoming edges as opposed to outgoing edges. This is due to the fact that the direction of certain predicates is chosen rather arbitrarily [e.g., (\texttt{Brussels}, \texttt{isCapitalOf}, \texttt{Belgium}) vs. (\texttt{Belgium}, \texttt{hasCapital}, \texttt{Brussels})]. This also allows for nodes from \texttt{Entities} to be in positions different from the starting position within walks. \\

Several mechanisms are implemented to speed up the walk extraction: (i) SPARQL requests to find the next hop in walks can be bundled together to reduce overhead introduced by HTTP when a remote KG is used, (ii) multi-threading is enabled to parallelize the extraction of walks, and (iii) caching is implemented to avoid redundant requests.

\subsection{CI/CT/CD and Documentation}
\label{subsec:ci}
To facilitate contributions by the open-source community to our code repository, multiple mechanisms have been set up. First, Continuous Integration (CI), through the use of Github Actions\footnote{\url{https://github.com/features/actions}}, is implemented which makes sure that the merge of the work of several developers does not impact the release of a project. With each push to one of the branches, several checks are performed, such as checking whether any styling guidelines have been violated. Second, Continuous Delivery (CD) is guaranteed as the \texttt{main} is always supposed to be the stable branch for which the checks performed by the CI pass. Added to that, the use of \texttt{poetry}~\footnote{\url{https://python-poetry.org/}} as dependency manager helps to facilitate future releases of \texttt{pyRDF2Vec} to the PyPI platform. Finally, a Continuous Testing (CT) mechanism executes a battery of unit tests, using \texttt{pytest}~\footnote{\url{www.pytest.org}}, for every push to the code repository. Afterwards, a coverage report is generated. With the help of these continuous methods, \texttt{pyRDF2Vec} has been able to
release several new features and fix bugs to increase its stability, popularity,
and notoriety. \\

Having an updated and clear documentation is essential for the proper use of a library and its evolution. Good documentation will make it easier to use and contribute to a library. To improve the clarity of the documentation in Python, \texttt{mypy}~\footnote{\url{http://mypy-lang.org/}}, an optional static type checker, can also be used in addition to PyDoc. While Python is natively a dynamically typed language, the use of such a static type checker requires that consistent types are filled in, which improved documentation. Finally, this documentation generation is done with Sphinx~\footnote{\url{https://www.sphinx-doc.org/}} and is automatically updated on the online website hosted by Read the Docs, at each commit on the \texttt{main} branch.

\section{Package Usage}\label{sec:usage}
\begin{figure}[h!]
    \centering
    \includegraphics[width=\textwidth]{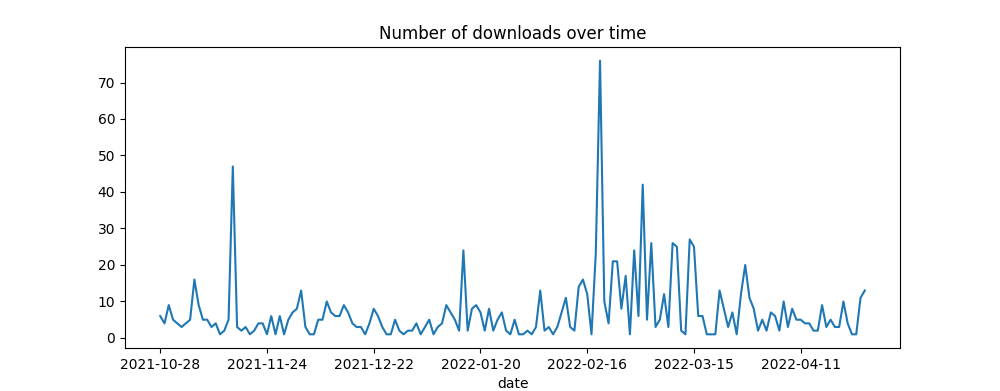}
    \caption{The number of downloads of the last 180 days of our \texttt{pyRDF2Vec} package.}
    \label{fig:downloads}
\end{figure}

At the time of writing, \texttt{pyRDF2Vec} has amassed 146 stars on Github and 17,500 downloads according to PePy\footnote{\url{https://pepy.tech/project/pyRDF2Vec}}. An overview of the number of downloads for the latest six months can be found in Figure~\ref{fig:downloads}. \\

\texttt{pyRDF2Vec} has been used in several research projects and practical use cases. As of today, \texttt{pyRDF2Vec} appears in 31 studies published on Google Scholar\footnote{\url{https://scholar.google.com/scholar?q="pyRDF2Vec"}}. We now give a brief overview of these studies. Ontowalk2vec~\cite{gkotse2022ontology} and Owl2Vec*~\cite{chen2021owl2vec} extend \texttt{pyRDF2Vec} to embed concepts by extracting walks from ontology information. Iana et al.~\cite{iana2020more} showed that applying reasoning to infer extra information in the KG before extracting walks results in little to no increased predictive performance. Portisch et al.~\cite{portischknowledge} compared embedding techniques suited for link prediction and suited for data mining on both link prediction and data mining tasks. \texttt{pyRDF2Vec} was used as one of the data mining techniques during evaluation. In \cite{jain2021embeddings}, \texttt{pyRDF2Vec} among many other embedding techniques, has been compared to non-embedding methods to better understand their semantic capabilities. In Sousa et al.~\cite{sousa2021supervised} \texttt{pyRDF2Vec} is used to tailor aspect-oriented semantic similarity measures to fit a particular view on biological similarity or relatedness in protein-protein, protein function similarity, protein sequence similarity and phenotype-based gene similarity tasks. Engleitner et al.~\cite{engleitnerknowledge} compare \texttt{pyRDF2Vec} with other embedding techniques for news article tag recommendation. Shi et al.~\cite{shi2021keyword,shi2021efficient} use \texttt{pyRDF2Vec} to calculate semantic similarity between concepts in several datasets. Gurbuz et al.~\cite{gurbuz2022knowledge} evaluate many different techniques, including \texttt{pyRDF2Vec}, for explainable target-disease link prediction. Steenwinckel et al.~\cite{steenwinckel2022ink} compare their newly proposed technique, INK, to state-of-the-art techniques such as \texttt{pyRDF2Vec}. Finally, Degraeve et al.~\cite{degraeve2022r} qualitatively compare embeddings produced by \texttt{pyRDF2Vec} with embeddings produced by their proposed RR-GCN through a t-SNE plot.

\section{Conclusion and Future Work}\label{sec:conclusion}

This paper presented the \texttt{pyRDF2Vec} software package. It reimplements the well-known RDF2Vec algorithm in Python, as this language is several significantly more popular in the data science community than Java, in which RDF2Vec was originally implemented. This reimplementation allows for data scientists to integrate RDF2Vec immediately into their pipeline. In addition to the original algorithm, \texttt{pyRDF2Vec} implements many extensions that have already been published, provides additional information and speeds up the walk extraction. The fact that these extensions are bundled in a single place could facilitate future research. The \texttt{pyRDF2Vec} architecture is set up in such a way, in combination with automatic styling, testing, and documentation to foster future external contributions. Several research projects and use cases have already used \texttt{pyRDF2Vec} in their experimentation or as a basis for their code, which we discuss in this paper.  \\

\paragraph*{Resource Availability Statement:} \texttt{pyRDF2Vec} is available under a MIT license on Github\footnote{\url{https://github.com/IBCNServices/pyRDF2Vec}}.

\section*{Acknowledgements}
Bram Steenwinckel (1SA0219N) is funded by a strategic base research Grant of the Fund for Scientific Research Flanders (FWO).

\bibliographystyle{splncs04} 
\bibliography{main}

\newpage
\appendix
\section{Appendix: Example Usage}\label{appendix}
We now provide a simple code snippet in Listing~\ref{code} that demonstrates how a user can generate embeddings for nodes of interest in his/her KG with just a few lines of code.

\begin{code}
\small
\captionof{listing}{Example usage of \texttt{pyRDF2Vec}}
\label{code}
\vspace{-1.5em}
\begin{minted}[mathescape,
               linenos,
               numbersep=5pt,
               frame=lines,
               framesep=2mm]{Python}
# entities is a list of URIs which we want to embed.
entities = [ ... ]

# Loads a KG object from hard disk, removes triples with 
# "dl#isMutagenic" as predicate, and specifies the paths 
# where literals can be found.
dl = "http://dl-learner.org/carcinogenesis"
kg = KG(
    "mutag.owl",
    skip_predicates={dl + "#isMutagenic"},
    literals=[
        [
            dl + "#hasBond",
            dl + "#inBond",
        ],
        [
            dl + "#hasAtom",
            dl + "#charge",
        ],
    ]
)

# Create a Word2Vec embedder that trains for ten epochs.
embedder = Word2Vec(workers=1, epochs=10)

# Create a Sampler that uses PageRank (damping 0.85).
sampler = PageRankSampler(alpha=0.85)

# Use HALK strategy to extract all walks of depth 2.
walker1 = HALKWalker(2, None, n_jobs=4, sampler=None)

# Create walker that samples 100 walks per entity.
walker2 = RandomWalker(2, 100, n_jobs=4, sampler=sampler)

# Create our transformer object.
transformer = RDF2VecTransformer(
    embedder,
    walkers=[walker1, walker2]
)

# Extract the embeddings and literals.
embeddings, literals = transformer.fit_transform(kg, entities)

\end{minted}
\end{code}

\end{document}